\documentclass[preprint,12pt,nopreprintline]{elsarticle}

\usepackage{framed,multirow}
\usepackage{amssymb}
\usepackage{latexsym}
\usepackage{lineno}
\usepackage{url}

\usepackage{autobreak} 
\usepackage[utf8]{inputenc} 
\usepackage[T1]{fontenc}    
\usepackage{hyperref}       
\usepackage{amsfonts}       
\usepackage{url}            
\usepackage{nicefrac}       
\usepackage{cite}
\usepackage{amsmath}
\usepackage{graphicx}
\usepackage{amssymb}
\usepackage{textcomp}
\usepackage{booktabs}
\usepackage{adjustbox}

\newcommand{\param}{\mathbf{m}}
\newcommand{\params}{\mathbf{u}}
\newcommand{\paramt}{\boldsymbol{\theta}}
\newcommand{\shape}{\mathcal{S}}
\newcommand{\speed}{\mathbf{v}}
\newcommand{\surface}{\shape}
\newcommand{\curve}{\mathcal{C}}
\newcommand{\euc}{\mathbb{R}}

\newcommand{\pos}{\mathbf{p}}

\newcommand{\normal}{\mathbf{n}}
\newcommand{\Identity}{\mathbf{I}}
\newcommand{\matSup}{\mathbf{M}}
\newcommand{\matInf}{\mathbf{N}}
\newcommand{\matB}{\mathbf{B}}
\newcommand{\matC}{\mathbf{C}}
\newcommand{\ex}{\mathbf{e}_x}
\newcommand{\ey}{\mathbf{e}_y}
\newcommand{\ez}{\mathbf{e}_z}
\newcommand{\tors}{\mathbf{t}}
\newcommand{\rot}{\mathbf{r}}

\newcommand{\distsq}{d_i^2(\mathbf{m})}

\begin{document}



\title{Second Order Kinematic Surface Fitting in Anatomical Structures} 

\author[1]{Wilhelm {Wimmer}\corref{cor1}}
\cortext[cor1]{Corresponding author at: Department of Otorhinolaryngology, Technical University of Munich, Trogerstr. 32, 81675 Munich, Germany\\ \textit{E-mail:} wilhelm.wimmer@tum.de}
\author[2]{Herv\'{e} {Delingette}}

\address[1]{Technical University of Munich, Germany; TUM School of Medicine, Klinikum rechts der Isar, Department of Otorhinolaryngology} 
\address[2]{Universit\'{e} C\^{o}te d'Azur, France; Inria Sophia Antipolis, Epione}


\begin{abstract}
Symmetry detection and morphological classification of anatomical structures play pivotal roles in medical image analysis. 
The application of kinematic surface fitting, a method for characterizing shapes through parametric stationary velocity fields, has shown promising results in computer vision and computer-aided design. 
However, existing research has predominantly focused on first order rotational velocity fields, which may not adequately capture the intricate curved and twisted nature of anatomical structures. 
To address this limitation, we propose an innovative approach utilizing a second order velocity field for kinematic surface fitting. This advancement accommodates higher rotational shape complexity and improves the accuracy of symmetry detection in anatomical structures. 
We introduce a robust fitting technique and validate its performance through testing on synthetic shapes and real anatomical structures. 
Our method not only enables the detection of curved rotational symmetries ({\em core lines}) but also facilitates morphological classification by deriving intrinsic shape parameters related to curvature and torsion. 
We illustrate the usefulness of our technique by categorizing the shape of human cochleae in terms of the intrinsic velocity field parameters.
The results showcase the potential of our method as a valuable tool for medical image analysis, contributing to the assessment of complex anatomical shapes.
\end{abstract}

 
\maketitle

\section{Introduction}
Symmetry is a fundamental concept in mathematics, shape analysis, and various scientific disciplines.
It refers to the property of a feature (points, lines, surfaces, or other mathematical objects) that remains unchanged under specific transformations, such as reflections, rotations, translations, or combinations of these operations. 
Symmetry plays a crucial role in understanding the organization and order found in the world, and is widely observed in both biological and non-biological systems.

For biomedical purposes, geometrical symmetry becomes particularly significant. 
Anatomical structures in the human body often exhibit symmetrical properties.
Detecting and analyzing symmetrical features can assist in the classification of the structure into a known taxonomy, identifying abnormalities, assessing asymmetries caused by injuries, diseases, or developmental issues, and aiding in surgical planning or implant design, e.g., for the cardiovascular system, bones, or the inner ear \citep{wimmer2019,rueckert1997automatic,eckhoff2016bilateral}.

Kinematic surface fitting enables to extract stationary velocity fields from oriented point clouds \citep{Hofer2005,Andrews2013}.
For shape analysis, the method offers two interesting features.
First, stationary velocity fields can be expressed as a compact set of parameters, allowing the global characterization of geometries analogous to primitive surfaces such as spheres, cylinders, cones, helices, conical spirals, and other surfaces of revolution. 
Second, stationary velocity fields can be used to compute critical points (e.g., zero-velocity convergence centers) and extract invariants to detect translational or rotational symmetry in meshes.

In the field of medical image analysis, kinematic surface fitting has been of limited use.
Since the velocity fields studied contain constant rotational components, they can only be applied to shapes with straight rotationally symmetric axes \citep{Andrews2013,Hofer2005} or used to approximate more complex structures \citep{wimmer2019}.
In contrast, naturally evolved structures often have shapes with curved rotational symmetry axes or {\em core lines}.
An illustrative example is the human cochlea, the organ of hearing.
The spiral shape consists of successive turns that are not parallel, but twisted against each other \citep{shin2013quantitative}.

Motivated by the goal of extracting and classifying intrinsic shape properties of the human cochlea \citep{wimmer2019human}, here we investigate the applicability of kinematic surface fitting using a nonlinear velocity field.
First, we provide a brief overview of conventional (i.e., first order) kinematic surface fitting.
Then, we derive a robust fitting scheme for a second order  velocity field and address the identification of core lines and critical points.
In the final section of the paper, we discuss the capabilities and limitations of the method using selected examples of test geometries and anatomical structures.

The paper presents several novel contributions.
First, we introduce second order velocity fields to the field of kinematic surface fitting. 
These velocity fields allow for the representation of more complex shapes, including those with curved rotational axes or core lines.
Second, we provide a robust fitting method for the proposed second order velocity field that also accounts for outliers using the heavy tailed Student-t distribution.  
Third, we demonstrate the applicability of our method to extracting convergence points and core lines in anatomical structures, including the aorta and the cochlea.
Forth, we illustrate how the method can be used for morphological classification of anatomical structures based on global intrinsic shape properties.  

The presented work extends our methods introduced  in a conference paper~\citet{wimmer2019}, which applied kinematic surface fitting using a linear velocity field to detect a straight rotational symmetry axis in human cochleae. This paper presents in more details the concept of kinematic surfaces and its links with other concepts in computational anatomy. Besides, we 
introduce  a non-linear velocity field (i.e., quadratic with respect to position) for kinematic surface fitting which remains
a linear function of its parameters. 
Finally,  we demonstrate the first application of morphometric classification of human cochlea based on intrinsic global shape properties derived from the parameters of that velocity field.

\section{Kinematic Surfaces}
Our work aims at extracting global geometric characteristics of anatomical structures by estimating the parameters of velocity fields that can be associated with those structures when considered as kinematic surfaces. 
In this section, we introduce the concept of kinematic surfaces, and describe their geometric properties based on the analysis of the velocity fields.

\subsection{Definition}
A kinematic surface is a surface which is at every point tangent to a parameterizable  velocity field.
We define a 3-dimensional parametric stationary velocity field $\speed(\pos,\param)\in\euc^3$ with $\param$ being a compact set of parameters and $\pos\in\euc^3$ being a point in 3-dimensional space.
A smooth surface $\surface\in\euc^3$ is a kinematic surface if it is tangential to a stationary velocity field $\speed(\pos,\param)$, i.e., if for each point lying on the surface $\pos\in\surface$ having a normal vector $\normal(\pos)$, the velocity field is in the tangent plane of the surface at $\pos$: $\speed(\pos,\param)\cdot\normal(\pos)=0$. 

The vector field $\speed(\pos,\param)$ is a tangent field of a surface, but it is also a velocity field, since a curve $\curve(u)$ can be grown from any seed point $\pos_\mathrm{s}$ lying on a surface by solving the differential equation $\frac{\partial \curve}{\partial u}=\speed(\curve(u))$, with $\curve(0)=\pos_\mathrm{s}$. 
This curve generated from $\pos_\mathrm{s}$ is likely to lie on the surface $\surface$ unless $\pos_\mathrm{s}$ is on the border of $\surface$. 

\subsection{Link with previous work}
\paragraph{Symmetric Surfaces}
The notion of kinematic surfaces is closely linked with the notion of group symmetry in 3-dimensional surfaces.  
Indeed, a parametric surface $\surface(\params)\in\euc^3$, $\params\in\Omega_S$ is considered to be symmetric if it is invariant by the application of a transformation (considered to be a diffeomorphism) of the 3D Euclidean space. 
More formally, if we write  $T(\pos;\paramt): \euc^3\longrightarrow\euc^3$, a symmetry transformation parameterized by a parameter $\paramt$, then $\forall \params\in\Omega_S, T(S(\params);\paramt)\in S$.  
The invariance can be written more explicitly as $T(S(\params);\paramt)=S(\params^\star(\params,\paramt))$ where $\params^\star(\params,\paramt)$ is the mapping associating the parameter $\params$  with the parameter of its symmetric point.

The notion of kinematic surfaces is related to that  of \emph{continuous symmetry} where there is a one dimensional continuous function $\paramt(t)$, $t\in[0,b]\in\euc$ for which the symmetry is defined. 
This implies that  $T(S(\params);\paramt(t))=S(\params^\star(\params,\paramt(t)))$, $\forall t\in[0,b]$ and $\params^\star(\params,\paramt(0))=\params$. 
In other words, there is a continuous set of symmetric transformations that maps each point to a curve on the surface which starts from that point. 
Then we can consider the derivative of the symmetric transformation at $t=0$, $\frac{\partial T}{\partial t}(S(\params);\paramt(0))=\frac{\partial S}{\partial t}(\params^\star(\params,\paramt(0)))$ and show that this is a vector field lying on the tangent plane of the surface at $S(\params)$. 
Conversely, given a stationary velocity field and a kinematic surface, it is easy to define a set of symmetric transformations by integrating the velocity field over a fixed amount of time. 

The concept of continuous symmetry can be opposed to that of discrete symmetry such as plane or rotational symmetries, where there exists only a discrete set of symmetric transformations for which the surface is invariant. 
Besides, continuous symmetry is related to key concepts in fundamental physics (like the definition of the action of a dynamic system) and to the notion of one-parameter subgroups of Lie groups in mathematics, both of which fall outside the scope of this paper.  

\paragraph{Symmetry Detection in Anatomical Structures} Detection of global plane symmetries on anatomical structures such as the brain has been explored in previous work ~\citep{liu2009symmetry}. 
The proposed methods can be posed as robust optimization problems \citep{thirion2000statistical}, as the local estimation of texture fractal dimensions~\citep{jayasuriya2013brain} or by using convolutional neural networks~\citep{9879810,8917648}. 
The estimation of the long axis of the left ventricle has also been achieved in various ways, including fitting a line through the circle centers detected with Hough transform~\citep{VANSTRALEN2008196}. 
Centerline extraction of blood vessels is often based on the local multiscale filtering~\citep{sironi2015multiscale} of the image, revealing the axial symmetry of the structure. 

Thus, prior work attempts to extract either global simple primitives (planes or lines) of symmetry or non-parametric local centers of symmetry. 
The method proposed in this paper, based on kinematic surfaces, has a complementary objective by extracting global non-trivial shapes primitives in a robust way.

\paragraph{Parametric Stationary Velocity Field for Image Registration} Stationary velocity fields have been introduced to parameterize diffeomorphisms for solving image registration problems~\citep{vercauteren2008symmetric}. 
However, those velocity fields often consist of multidimensional images where a speed vector is stored at each voxel. 
In the proposed method, we consider very compact parametric fields consisting of a dozen parameters instead of millions. 
\citet{arsigny2009fast} have, however, introduced polyaffine registration approaches leading to a combination of affine registrations.

This paper considers the motion invariance of shapes with a second order velocity field that generalizes a single affine transformation.

\paragraph{Computation Fluid Dynamics (CFD) Flow Analysis} 
The CFD community has developed a large body of techniques for visualizing flow fields~\citep{merzkirch2012flow,Gunther2018} and extracting quantitative features (e.g., vortices and vortex core lines) from those flow fields. 
Equivalent problems arise when analyzing velocity fields associated with kinematic surfaces, especially when dealing with higher order velocity fields. 
Therefore, the same tools can be applied for both domains. 

\subsection{Properties of Kinematic Surfaces}
\label{subsec;propKinematicSurf}
Given a kinematic surface $\surface$ and its corresponding velocity field $\speed(\pos,\param)$, the parameter set $\param$ characterizes the shape of the surface. However, this parameter set is not unique since any co-linear velocity field $\mu(\pos) \speed(\pos,\param)$, $\mu(\pos)\in\euc$ is also a valid velocity field for the same kinematic surface. 
Furthermore, the geometric interpretation of the parameter set $\param$ may not be obvious, as seen for the higher order velocity field.

The analysis of the parametric velocity field $\speed(\pos,\param)$ provides a framework for extracting meaningful geometric features.  For  instance, \emph{points of convergence} (i.e., critical points) are points where the velocity is null: $\speed(\pos_0)=0$. These are points of accumulation or repulsion, depending on the sign of the velocity divergence $\nabla_p\cdot\speed(\pos_0,\param)$ at this location. 
The nature of the convergence points can be analyzed by looking at the eigenvalues (that can be complex or real) of the Jacobian matrix $\mathbf{J}(\pos)=\nabla^T_p \speed(\pos,\param)$  \citep{35197}.

Furthermore, an \emph{axis of symmetry} defined by a point $\mathbf{d}_0$ and a direction vector $\mathbf{d}$ corresponds to a straight line, where the velocity is directed along the direction $\mathbf{d}$ of this line: $\speed(\mathbf{d}_0+\lambda\mathbf{d})\times\mathbf{d}=\mathbf{0}$. Any seed point located in this axis will remain on this axis when moved according to the velocity field.

A generalized notion of curve of symmetry is given by \emph{core lines} (corresponding to vortex core lines in the CFD community). 
There exists many competing definitions of vortices and vortex core lines \citep{Post2003}, but we rely on the approach proposed by \citet{roth1998higher} that defines those lines as the points where the torsion of streamlines is null (i.e., where the derivative of the acceleration is parallel to the velocity field). 
Besides, it is possible to eliminate weak or short vortices by thresholding on two local attributes. 
We use the parallel vector operator defined in \citet{van2009using} for a practical implementation.
The detected core lines correspond geometrically to a symmetry curve around which the surface is wrapped.

\subsection{First Order Kinematic Surfaces}

To the best of our knowledge, the concept of kinematic surfaces has only been developed so far with linear or first order velocity fields \citep{Andrews2013}, i.e., fields for which the velocity is a linear function of the position: $\speed(\pos,\param)=\mathbf{A}\pos+\mathbf{c}$, where $\mathbf{A}$ is a $3\times 3$ matrix and $\mathbf{c}$ a 3-dimensional vector.  The considered matrix $\mathbf{A}$ is restricted to be the sum of an antisymmetric matrix and a matrix proportional to the identity matrix, such that:
\begin{equation}
    \speed(\pos,\param)=\mathbf{A}\pos+\mathbf{c}= \rot\times\pos +\mathbf{c} +\gamma\pos ,
    \label{eq;first order}
\end{equation}

where $\param=[\rot,\mathbf{c},\gamma]$.
Those linear velocity fields lead to a large family of kinematic surfaces that have simple geometric characteristics. 
We detail below some cases associated with this field for increasing levels of complexity.

\paragraph{Constant Velocity Field}
The constant field $\speed(\pos,\param) = \mathbf{c} $ only allows translational motions along $\mathbf{c}$ and describes a linear extrusion of a generator curve.

\paragraph{Scale Velocity Field}
The scale field $\speed(\pos,\param) = \mathbf{c} +\gamma\pos $ also includes scaling by factor $\gamma$.

\paragraph{Rotational Velocity Field}
The first order rotational field $\speed(\pos,\param) = \rot\times\pos + \mathbf{c}$ 
contains a rotational axis along the vector $\rot$ and can be used to express symmetric rotational surfaces or a straight helix.

\paragraph{Rotational Scale Velocity Field}
The first order rotational scale field can be expressed as $\speed(\pos,\param) = \rot\times\pos +\gamma\pos  +\mathbf{c} $, encoding cylindrical, conical, rotational, and helical motions, i.e., motions composed of rotation $\rot$, translation $\mathbf{c}$ and scaling $\gamma$ \citep{Hofer2005}. 
Since scaling is contained within the magnitude of the rotation vector, effectively, this velocity field has only one true degree of freedom to describe spiral shape invariance \citep{Pottmann2006}.

\paragraph{Convergence Point}
We can find the convergence center $\pos_0$ by solving $\speed(\pos_0,\param)\\=0$ \citep{Hofer2005}:

\begin{equation}
    \pos_0 = \frac{1}{\gamma ( \mathbf{r}^{2} + \gamma^{2})} \Big(\gamma \mathbf{r} \times \mathbf{c} -\gamma^{2}\mathbf{c}-(\mathbf{r} \cdot \mathbf{c})\mathbf{r}\Big).
\end{equation}

The velocity field can then be alternatively expressed with respect to $\pos_0$:

\begin{equation}
\label{eq:alter_velo}
    \speed(\pos,\param)=\rot\times(\pos-\pos_0)+\gamma(\pos-\pos_0).
\end{equation}

\paragraph{Streamlines}
The continuous symmetry is defined by the streamlines $\curve(u)$ defined by the integration of the velocity $\frac{\partial \curve}{\partial u}=\speed(\curve(u))$. 
With first order velocity field, such streamlines are {\em conical-spiral} ({\em aka} concho-spiral) curves and can be computed in closed form as: 
\[
\pos(u)=\pos_0+a  \exp(\gamma u)\left ( \cos(r (u-u_0) \ex + \sin(r (u-u_0) \ey + b\ez \right )
\]  where $u_0,a,b$ are any scalar and $\rot= r \ez$ and $(\ex,\ey,\ez)$ make an orthonormal frame.

\paragraph{Symmetry Axis} With the velocity of Eq.\ref{eq;first order}, the straight line passing through $\pos_0$ and directed by $\rot$ is an axis of symmetry around which each streamline turns around.

\section{Gaussian Kinematic Surface Fitting}
\label{sec;fitting}
\subsection{General Approach}
We consider the problem of fitting a kinematic surface given  a set of points with position $\pos_i$ and their unit surface normal $\normal_i$. 
This implies finding a parametric  velocity field $\speed(\pos,\param)$ for which the velocity field lies in the tangent plane at each sample point of the discrete surface $\mathcal{S}=\{\pos_i,\normal_i\}$:
\begin{equation}
    \speed(\pos_i,\param) \cdot \normal_i = 0,~~~\forall i
\end{equation}

\citet{Andrews2013} provide a generalized and basis-independent formulation of the kinematic surface fitting problem using an approximate maximum likelihood approach \citep{Chernov2007}. With the proposed framework,  the optimal parameter vector $\param$ is the  one minimizing  the following distance $d_i(\mathbf{m)}$~:
\begin{equation}
\label{eq:distance_fit}
    d_i(\param)=\frac{\speed(\pos_i,\param)\cdot\normal_i}{\sqrt{\|\speed(\pos_i,\param)\|^2+w_\pos\|\nabla_\pos(\speed(\pos_i,\param)\cdot\normal_i)\|^2}},
\end{equation}
where $w_\pos$ is a scalar. 
This distance measures how well the normalized velocity field lies in the tangent plane. 
The velocity normalization avoids selecting the trivial solution $\speed(\pos,\param)=\mathbf{0}$ and is composed of 2 terms: the velocity norm and the norm of the gradient of the velocity-normal dot product. 
This second term scaled by $w_\pos$ aims  to control the gradient norm of the distance (somewhat enforcing its Lipschitz continuity), thus to make the square distance as convex as possible. 

We then assume that for a set of oriented points, the distance follows a Gaussian distribution  with variance $\Sigma$, that is $p(d_i|\param)={\cal N}(d_i|0,\Sigma)$. Then, we want to determine $\param$ to maximize the log-likelihood:
\begin{align*}
\label{eq:distance_fit}
    \log p(\mathcal{S}|\param)&=\log \prod_{i=1}^n p(d_i|\param)=  
    -\frac{n}{2}\log{2\pi \Sigma} - \frac{1}{2} \sum_{i=1}^n \frac{\distsq}{\Sigma} \\
    &=-\frac{n}{2}\log{2\pi \Sigma} + \mathcal{L}(\param).
\end{align*}
Note that it is not necessary to provide a consistent orientation of the points $\pos_i$ since the criterion is invariant to a change of orientation when replacing $\normal_i$ with $-\normal_i$.

\citet{Andrews2013} make the additional assumption that the velocity field is a linear function of its parameters $\param$: $\frac{\partial^2 \speed(\pos,\param)}{\partial \param_j \param_k}=\mathbf{0}$. 
This implies that there exists a matrix $\mathbf{H}(\pos)$ such that $\speed(\pos,\param)=\mathbf{H}(\pos) \param$  and a vector $\mathbf{f}(\pos,\normal)$ such that  $\speed(\pos,\param)\cdot\normal=\param\cdot \mathbf{f}(\pos,\normal)$. As a consequence, the square distance $d_i^2$ is a ratio between two quadratic forms of $\param$:
\begin{equation}
\label{eq:squared_dist}
    \distsq=\frac{(\speed(\pos_i,\param)\cdot\normal_i)^2}{\|\speed(\pos_i,\param)\|^2+w_\pos\|\nabla_\pos(\speed(\pos_i,\param)\cdot\normal_i)\|^2}=\frac{\param^T\matSup_i\param}{\param^T\matInf_i\param}
\end{equation}
where $\matSup_i$ and $\matInf_i$ are two symmetric matrices.

Then, maximizing the log-likelihood $\mathcal{L}(\param)$ is equivalent to minimizing $\sum_{i=1}^n\frac{\param^T\matSup_i\param}{\param^T\matInf_i\param}$, which leads to the generalized eigenvalue problem \citep{Chernov2007}:
\begin{equation}
\label{eq:gen_eigen}
    \matB_m \param=\matC_m \param,
\end{equation} 
with $\matB_m=\sum_{i=1}^n \frac{\matSup_i}{\param^T\matInf_i\param}$ and $\matC_m=\sum_{i=1}^n \frac{\param^T\matSup_i\param}{(\param^T\matInf_i\param)^2}\matInf_i$. The non-linear problem can be solved by iteratively computing the matrices $\matB_m$ and $\matC_m$ for a given estimation of $\param$ and then estimating $\param$ as the eigenvector associated with the eigenvalue closest to zero.

\subsection{Fitting First Order Velocity Fields}
We consider the first order velocity field defined in Eq.\ref{eq;first order} where $\param=[ \mathbf{r},\mathbf{c},\gamma]$. It is easy to verify that the velocity field is a linear function of parameters $\param$.


To compute the matrices $\matSup_i$ and $\matInf_i$, we find that $\nabla_\mathbf{p}(\speed(\pos_i)\cdot\normal_i)=\normal_i \times \mathbf{r}+\gamma\normal_i=[\mathbf{A}_{\mathbf{r}}+\gamma\Identity]\normal_i$ where $\mathbf{A_{r}}$ is the skew-symmetric matrix associated with $\mathbf{r}$, and $\Identity$ is the identity matrix. 
Furthermore,  we have $\|\nabla_\mathbf{p}(\speed(\pos_i)\cdot\normal_i)\|^2=\|\normal_i \times \mathbf{r}\|^2+\gamma^2$ and $\|\speed(\pos_i)\|^2=\|\mathbf{r}\times\pos_i\|^2 +\gamma^2\|\pos_i\|^2   + 2 \gamma (\pos_i\cdot\mathbf{c})+2 [\mathbf{r},\pos_i,\mathbf{c}] + \|\mathbf{c}\|^2$. 
With a first order velocity field, the seven-dimensional vector $\mathbf{f}(\pos,\normal)$ is  $\mathbf{f}(\pos,\normal) := [\pos \times \normal, \normal, \pos \cdot \normal ]$.

To find the parameter vector $\param$, we solve equation (\ref{eq:gen_eigen}) after computing the $7\times 7$ matrices $\matSup_i$ and $\matInf_i$ \citep{Andrews2013}:
\begin{equation}
\label{eq:M} 
   \matSup_i=\mathbf{f}(\pos_i,\normal_i) \mathbf{f}(\pos_i,\normal_i)^T
\end{equation}
   
   and
   
\begin{equation}
   \matInf_i=\begin{bmatrix} \mathbf{A}_{\pos_i}^T\mathbf{A}_{\pos_i} + w_\pos \mathbf{A}_{\normal_i}^T\mathbf{A}_{\normal_i} & -\mathbf{A}_{\pos_i} & \mathbf{0}\\ 
   -\mathbf{A}^T_{\pos_i} & \Identity & \pos_i \\ 
   \mathbf{0} & \pos_i^T & \pos_i\cdot\pos_i +  w_\pos \end{bmatrix}. 
\end{equation}

\section{Second Order Velocity Field}
\subsection{Definition}

It is possible to identify a rotational symmetry axis of kinematic surfaces using a linear velocity field. 
However, for many anatomical structures, we hypothesize that such symmetry line may be curved and not straight. 
This is why we propose to consider a more sophisticated velocity field that is quadratic with respect to the position $\pos$: 

\begin{align}
\speed(\pos) = (\tors \times \pos)\times\pos + \rot \times \pos +\gamma\pos +\mathbf{c},
\end{align}
with the second order kinematic parameter $\tors$. 
With this parametric velocity field, we replace the fixed axis $\rot$ with the moving direction $\tors \times \pos+\rot$, which is rotating around the direction $\tors$. 
Besides, the parameter vector $\param=[\tors,\rot,\mathbf{c},\gamma]$ is now a 10-dimensional vector and the velocity field remains a linear function of the parameters in $\param$. 
Obviously, this second order velocity field generalizes the first order field, which corresponds to the case  $\tors=\mathbf{0}$.

\paragraph{Convergence Point}
There is no closed-form solution for isolating convergence points $\pos_0$ in the case of the second order velocity field. 
Therefore, to find the convergence point, we employ a nonlinear programming solver (“fminsearch”) minimizing the objective function defined as the norm of the velocity vector field ($\|\speed(\pos_i)\|$).
In analogy to Eq. \ref{eq:alter_velo}, we can find an alternative expression of the velocity field centered at $\pos_0$:
\begin{equation}
    \speed(\pos)=\Big( \tors\times(\pos-\pos_0) + \rot \Big) \times(\pos-\pos_0) +\gamma(\pos-\pos_0).
\end{equation}

\paragraph{Streamlines} The streamlines cannot be computed in closed form, but must be integrated through Euler or more sophisticated integration methods.

\paragraph{Core Line}
When $\tors\neq 0$, no straight axis of rotational symmetry can be found, but we can extract core lines as defined in section\ref{subsec;propKinematicSurf} that can be seen as curves of symmetry around which particles swirl. 
In practice, we follow vorticity extrema in the vector field, 
implemented using the parallel vector operator \citep{roth1998higher,van2009using}.

\subsection{Kinematic Surface Fitting}
 
To estimate the 10-dimensional vector $\param=[\tors,\rot,\mathbf{c},\gamma]$ from a set of oriented points, we proceed as described in Section~\ref{sec;fitting} since we have a linear relation between the velocity field and the parameter vector $\param$. 
For instance, the 10-dimensional vector $\mathbf{f}(\pos,\normal)$ now writes as 
$[ (\normal \times \pos) \times \pos, \pos \times \normal , \normal, \pos \cdot \normal ]$ and the matrix $\matSup_i$ can be obtained using Eq.~\ref{eq:M}.

To compute the normalization matrix $\matInf_i =\matInf^{(1)}_i + w_\pos \matInf^{(2)}_i$, we first express the term $\|\speed(\pos_i)\|^2$ as a multiplication of the matrix $\matInf_i^{(1)}$ and $\param$ where~:

\[
\matInf^{(1)}_i=
\begin{bmatrix} 
    \mathbf{F}_{\pos_i}^T \mathbf{F}_{\pos_i} & -\mathbf{F}_{\pos_i} \mathbf{A}_{\pos_i} & \mathbf{F}_{\pos_i}  & \mathbf{0}\\
    - \mathbf{A}_{\pos_i}^T \mathbf{F}_{\pos_i}^T & \mathbf{A}_{\pos_i}^T\mathbf{A}_{\pos_i} & -\mathbf{A}_{\pos_i} & \mathbf{0}\\ 
    \mathbf{F}_{\pos_i}^T  &   -\mathbf{A}^T_{\pos_i} & \Identity & \pos_i \\ 
    \mathbf{0} & \mathbf{0} & \pos_i^T & \pos_i\cdot\pos_i
\end{bmatrix} ,
\]
with $\mathbf{F}_{\pos_i} = \pos_i \otimes \pos_i -(\pos_i \cdot \pos_i) \Identity$, where $\otimes$ denotes the tensor product such that $\mathbf{a}\otimes\mathbf{b}=\mathbf{a}\mathbf{b}^T$. Moreover, we find that:
\begin{equation}
\label{eq:grad_vpn}
    \nabla_\mathbf{p}(\speed(\pos_i)\cdot\normal_i) =   (\tors \otimes \pos ) \normal + ( \tors \cdot \pos )  \normal - 2 (\pos \otimes \tors)  \normal + \normal \times \rot + \gamma \normal.
\end{equation} 

 Then, we can express $ \|\nabla_\mathbf{p}(\speed(\pos_i)\cdot\normal_i) \|^2$ in terms of the matrix $\matInf_i^{(2)}$ and $\param$:

\[
\matInf^{(2)}_i=
\begin{bmatrix} 
    \mathbf{G}_{\pos_i,\normal_i}^T \mathbf{G}_{\pos_i,\normal_i} & \mathbf{G}_{\pos_i,\normal_i} \mathbf{A}_{\normal_i} & \mathbf{0} & \mathbf{G}_{\pos_i,\normal_i} \normal_i\\
    \mathbf{A}_{\normal_i}^T \mathbf{G}_{\pos_i,\normal_i}^T & \mathbf{A}_{\normal_i}^T\mathbf{A}_{\normal_i}  & \mathbf{0} & \mathbf{0}\\ 
    \mathbf{0} &  \mathbf{0} & \mathbf{0} & \mathbf{0} \\
    \normal_i^T \mathbf{G}_{\pos_i,\normal_i}^T  &   \mathbf{0} & \mathbf{0} & \normal_i\cdot \normal_i= 1 
\end{bmatrix}, 
\]

with $\mathbf{G}_{\pos_i,\normal_i}  = (\pos_i \cdot \normal_i) \Identity -2 \pos_i \otimes \normal_i + \normal_i \otimes \pos_i$. As before, we obtain our parameters $\param$ by assembling matrices $\matB_m$ and $\matC_m$ and finally solving the generalized eigenvalue problem of Eq.~\ref{eq:gen_eigen}.

\section{Robust Kinematic Surface Fitting}
The method for estimating kinematic surface parameters in Section~\ref{sec;fitting} is a least square fitting method which is sensitive to outliers. Such outliers may come either from the erroneous estimation of surface normals $\normal_i$ at points $\pos_i$ or by including points that are not lying on the kinematic surfaces. Classical robust least square methods such as m-estimators or least trimmed squares\citep{lts} need to estimate precisely extra variables such as the percentage of outliers. 

 We adopt a parametric approach by replacing the Gaussian likelihood with a heavy tailed Student-t distribution. The benefit of this distribution is that Student-t is a Gaussian Scale Mixture which makes it amenable to a data-driven  iterative estimation with closed form solutions. 
 
More precisely, we now assume that $p(d_i|\param)={\mathrm{St}}(d_i|0,\Sigma,\nu) = \\ \int_{z_i}\mathcal{N}(d_i|0,\Sigma/z_i)\,\mathrm{Ga}(z_i|\nu/2,\nu/2) \mathrm{d}z_i$, where $z_i$ is the variance scale variable which has a prior given by the Gamma distribution parameterized by the degrees of freedom $\nu$. 
When $\nu\rightarrow +\infty$, the Student-t distribution is equivalent to the Gaussian distribution and the variable $\nu$ is inversely proportional to the number of outliers.

The robust estimation of a kinematic surface is now achieved with an Expectation-Maximization scheme, where $z_i$ is the latent variable \citep{Scheffler2008}, and where $\nu$, $z_i$ and $\Sigma$ are
iteratively estimated. 
In the E-step, we estimate the posterior distribution of $z_i$ as a Gamma distribution: $p(z_i|d_i,\nu,\Sigma)=\mathrm{Ga}(z_i|\frac{\nu+1}{2},\frac{\nu}{2}+\frac{\distsq}{2\Sigma})$.  
This implies that the mean value of the latent variable $z_i$ is $\frac{(\nu+1)}{(\nu+\distsq/\Sigma)}$. 
Thus, for large of $\nu$ (i.e. when the distribution is close to be Gaussian), $z_i$ approaches 1, irrespective of the  distance $\distsq$. 
Conversely, if $\nu$ is small, then $z_i$ is close to zero when the distance $\distsq$ is much larger than the variance $\Sigma$, i.e., when dealing with outliers.  
The quantity $\frac{\distsq}{\Sigma}$ acts here as a Mahalanobis distance.

In the M-step, the variance $\Sigma$ and degree of freedom $\nu$ are optimized with a fixed value of $z_i$ set to its mean value. The variance is then estimated as   $\Sigma=\frac{1}{n} \sum_{i=1}^n z_i \,\distsq$. 
The  value of  $\nu$ is not obtained in closed form, but is the solution of the following equation:
 \begin{align*}
     -\psi\Big(\frac{\nu}{2}\Big) + \log\Big(\frac{\nu}{2}\Big) + 1 +\psi\Big(\frac{\nu+1}{2}\Big) -\log\Big(\frac{\nu+1}{2}\Big)  +\frac{1}{n} \sum_{i=1}^n (\log z_i -z_i)=0,
\end{align*} 
where $\psi(x)$ denotes the digamma function. 
The value of $\nu$ after performing a few iterations using an implementation of the Levenberg-Marquardt algorithm \citep{liu1995ml}. 

Finally, after convergence of the EM algorithm, we get the estimation of the latent $z_i$ of each data point, which also indicates if that point is an inlier ($z_i\approx 1$) or an outlier ($z_i\approx 0$). 
The maximization of the log-likelihood with the Student-t distribution is then equivalent to solving equation (\ref{eq:gen_eigen}), but with $\matB_m=\sum_{i=1}^n z_i \frac{\matSup_i}{\param^T\matInf_i\param}$ and $\matC_m=\sum_{i=1}^n z_i \frac{\param^T\matSup_i\param}{(\param^T\matInf_i\param)^2}\matInf_i$. 
Thus, the robust estimation of kinematic surfaces involves two nested iterative processes, the inner one being the estimation of the variable $z_i$ softly characterizing each oriented point and the outer one estimating the velocity field parameter $\param$ with weighted contributions from each data point.

\section{Results}
\subsection{Validation}
\begin{figure}[ht!]
\centering
\includegraphics[trim= 130 40 90 50, clip,scale=.5]{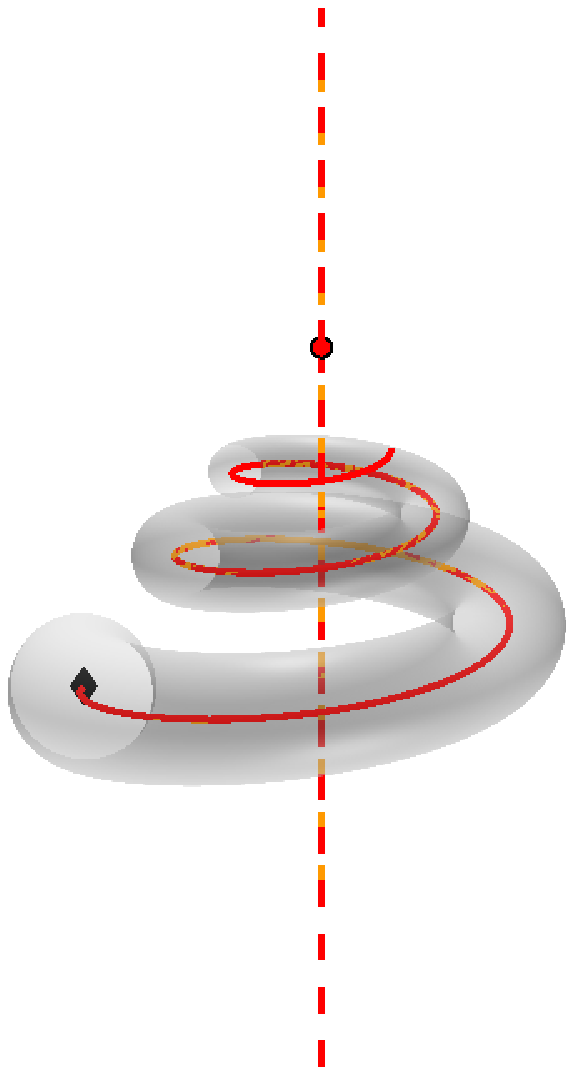}
\begin{picture}(0,0)\put(-110,100){\small{a}}\end{picture}
\includegraphics[trim=130 40 100 50, clip,scale=.5]{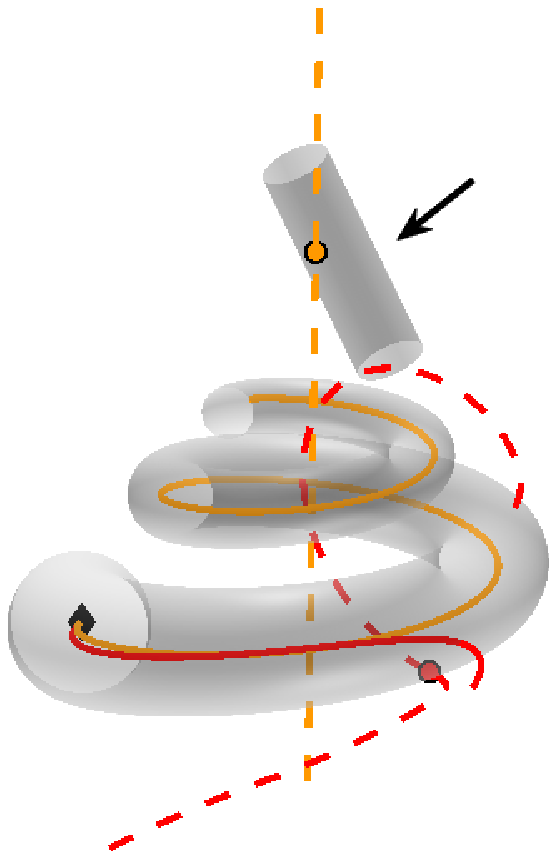}
\begin{picture}(0,0)\put(-100,100){\small{b}}\end{picture}
\includegraphics[trim=90 20 110 50, clip,scale=.5]{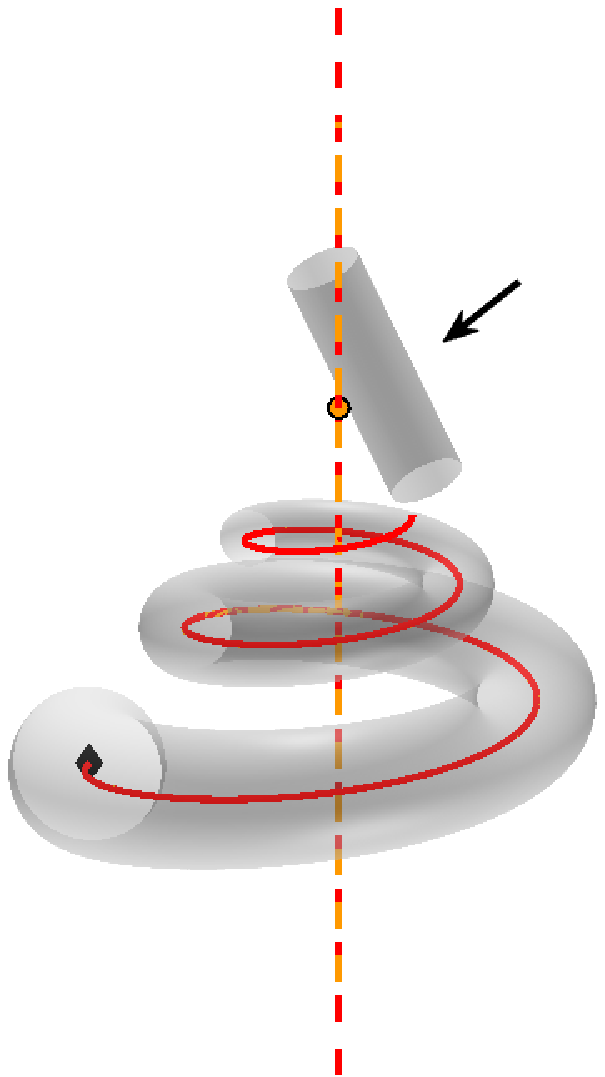}
\begin{picture}(0,0)\put(-80,100){\small{c}}\end{picture}
\includegraphics[trim= 70 20 10 0, clip,scale=.42]{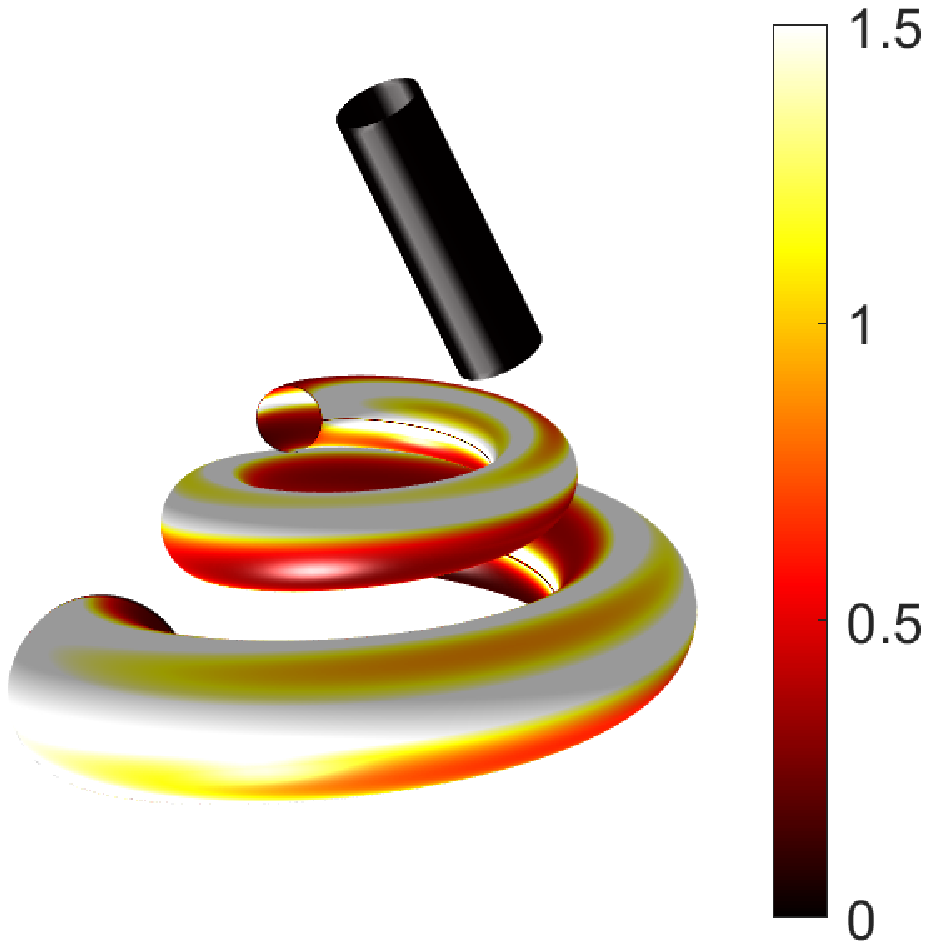}
\begin{picture}(0,0)\put(-110,100){\small{d}}\end{picture}
\caption{Kinematic surface fitting of a logarithmic spiral. Displayed are the results for the first order rotational (yellow curves) and second order rotational (red curves) velocity fields.
The dashed curves indicate the detected core lines, and convergence points $\pos_0$ are shown as circles.
Continuous curves represent example center lines $\curve$ traced from a central seed point $\pos_\mathrm{s}$ (black diamonds). a) Ideal case: both velocity fields converge to the same solution. b) Non-robust fitting with added cylindrical structure as outlier (arrow). c) Robust fitting of shape with outlier (arrow). d) Visualization of confidence weights $z_i$ estimated for robust fitting of the second order velocity field.} \label{fig:test_log_spiral}
\end{figure}

\begin{figure}[ht!]
\centering
\includegraphics[trim=90 10 30 0, clip,scale=.33]{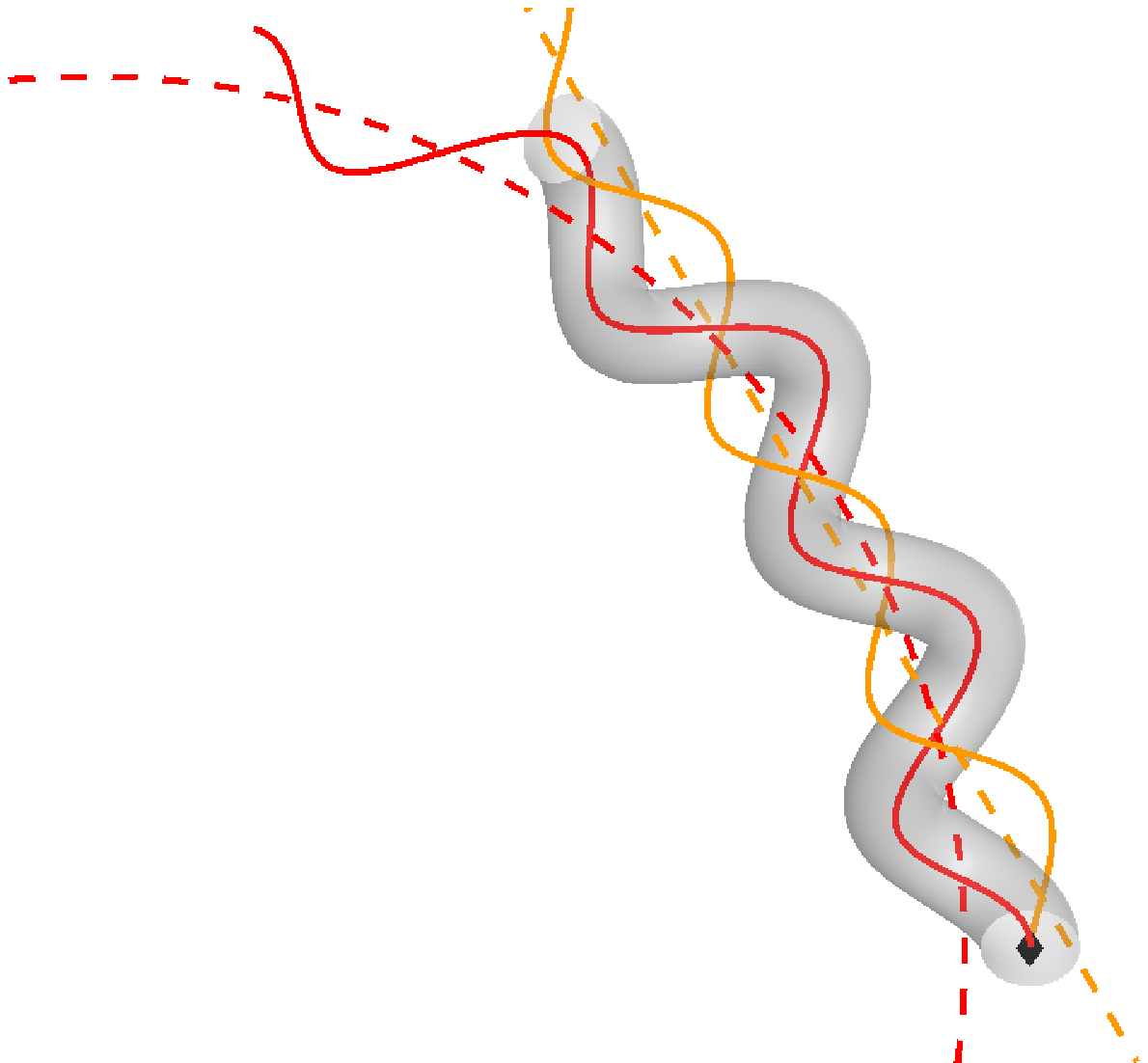}
\begin{picture}(0,0)\put(-120,100){\small{a}}\end{picture}
\includegraphics[trim=90 10 60 0, clip,scale=.37]{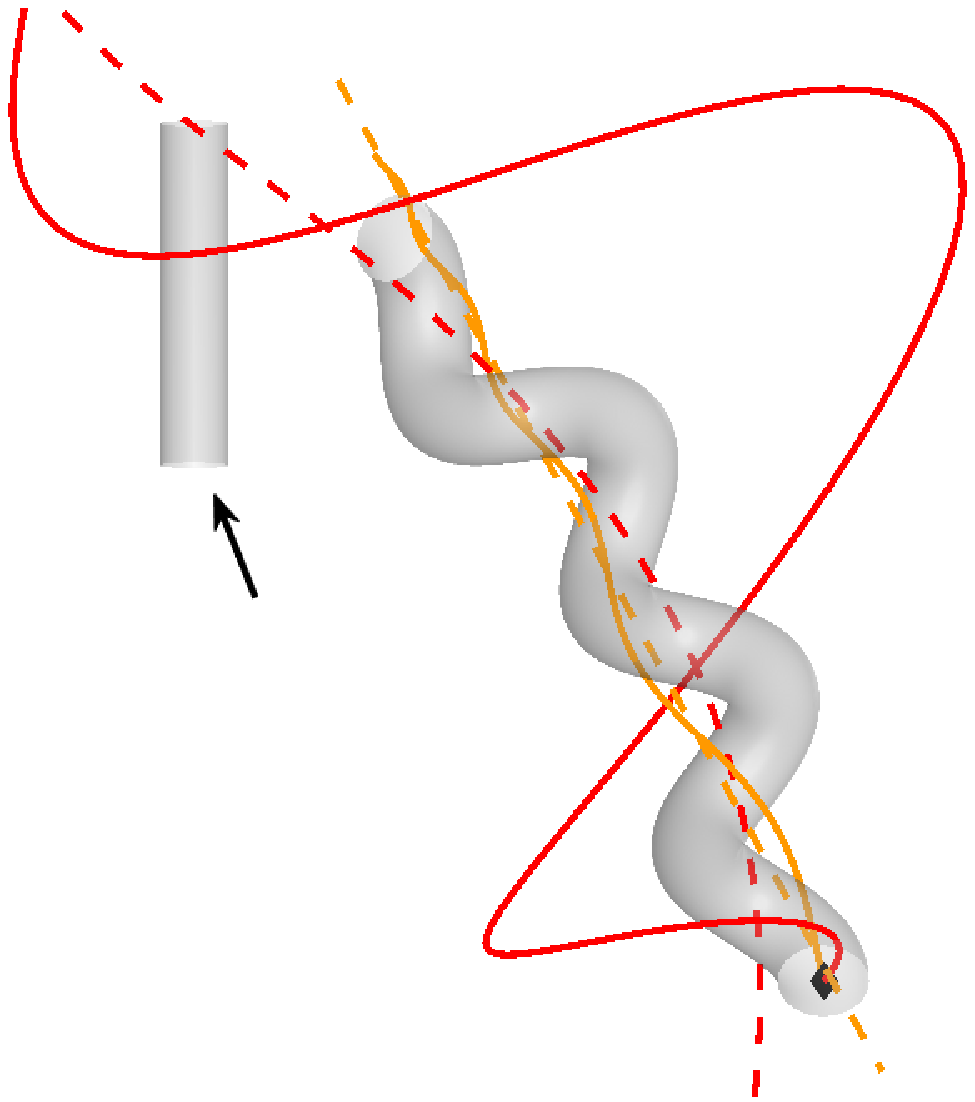}
\begin{picture}(0,0)\put(-120,100){\small{b}}\end{picture}
\includegraphics[trim=50 30 60 0, clip,scale=.4]{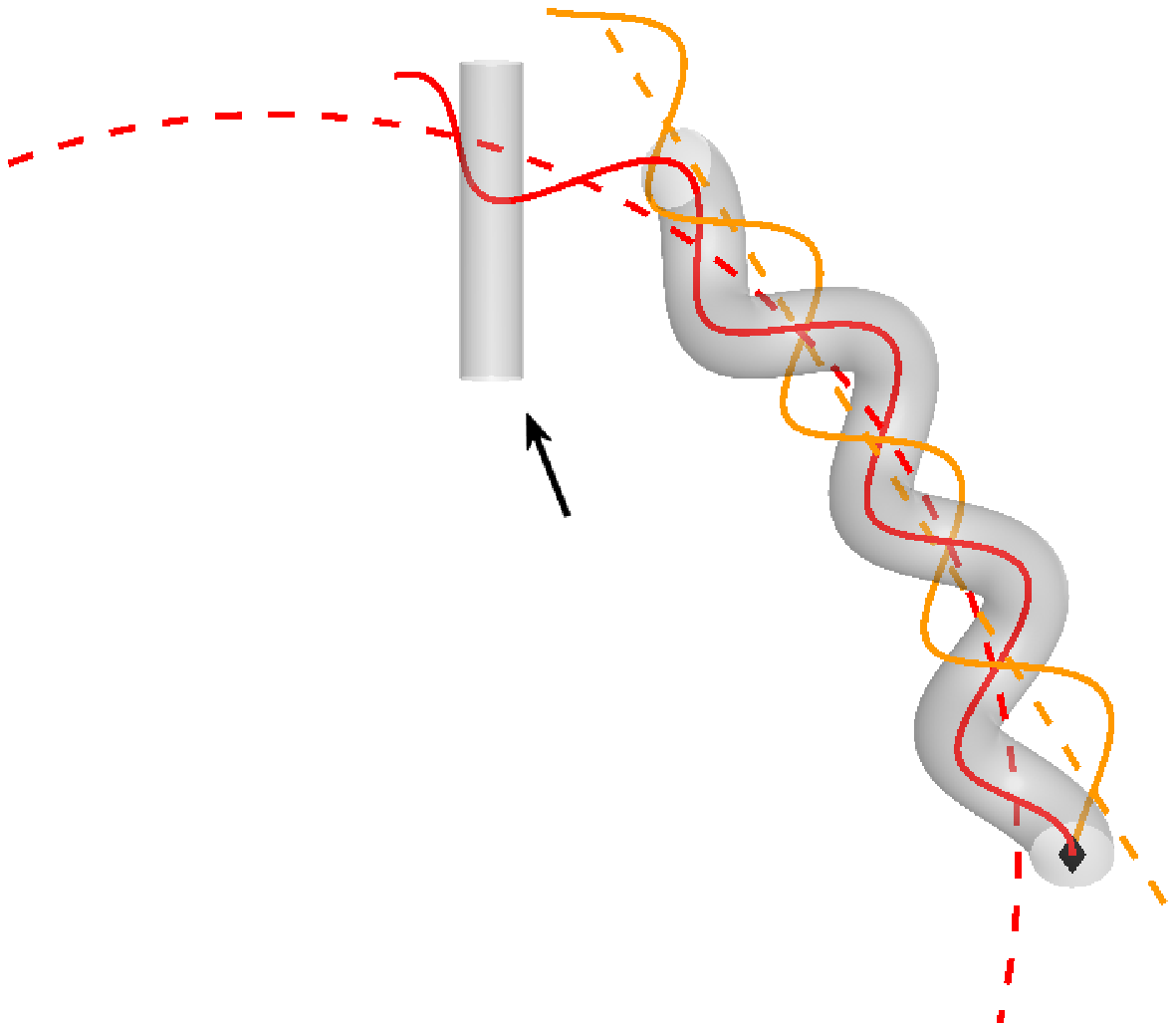}
\begin{picture}(0,0)\put(-120,100){\small{c}}\end{picture}
\includegraphics[trim=130 40 40 0, clip,scale=.45]{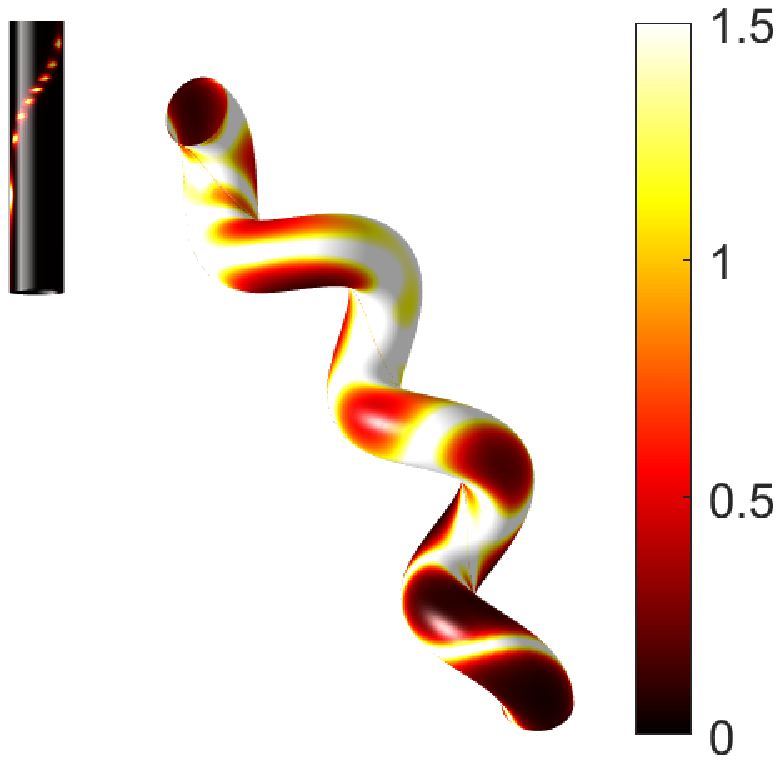}
\begin{picture}(0,0)\put(-120,100){\small{d}}\end{picture}
\caption{Kinematic surface fitting of a bent helix. Displayed are the results for the first order rotational (yellow curves) and second order rotational (red curves) velocity fields.
The dashed curves indicate the detected core lines. 
Continuous curves represent example center lines $\curve$ traced from a central seed point $\pos_\mathrm{s}$ (black diamonds). a) Ideal case: Second order fitting can capture the bending, in contrast to the first order approach. b) Non-robust fitting with added cylindrical structure as outlier (arrow): the second order fit does not converge properly. c) Robust fitting with outliers (arrow): the second order fitting method converges to the correct solution. d) Visualization of confidence weights $z_i$ estimated for robust fitting of the second order velocity field.} \label{fig:test_bent_helix}
\end{figure}

Figures \ref{fig:test_log_spiral} and \ref{fig:test_bent_helix} illustrate examples of non-robust and robust kinematic surface fitting in synthetic test shapes. 
The fitting results are listed in Table \ref{tab:validation}.
In the first example, a logarithmic helical spiral, the fit second order velocity field becomes degenerate, and is identical to the first order field solution (Fig. \ref{fig:test_log_spiral}a) as $\tors\approx 0$. 
Introducing outliers (cylindrical structure) distorts the solution of the second order velocity field (Fig. \ref{fig:test_log_spiral}b). 
However, it can be recovered with the robust fitting procedure (Fig. \ref{fig:test_log_spiral}c and Fig. \ref{fig:test_log_spiral}d).

The second example shows a curved helix. 
It is obvious that the first order velocity field is incapable of capturing the bent core line of the structure, while the second order velocity field is correctly detecting the curvature (Fig. \ref{fig:test_bent_helix}a). 
Adding outliers (cylindrical structure), results in a degraded solution of the second order velocity field fit (Fig. \ref{fig:test_bent_helix}b). 
The robust estimation scheme can identify the outliers in terms of a different kinematic and assigns low confidence to the data points of the outliers (Fig. \ref{fig:test_bent_helix}c and Fig. \ref{fig:test_bent_helix}d).

\begin{table*} 
\centering
\caption{Validation of fitting on test surfaces (15 iterations, $w_\pos = 0.001$).}
  \begin{adjustbox}{width=0.85\textwidth}
\begin{tabular}{lcccccccc}
\toprule
 & \multicolumn{2}{c}{\textbf{First order rotational scale velocity field}}  & & \multicolumn{2}{c}{\textbf{Second order rotational scale velocity field}}  \\
\cmidrule(lr){2-3}\cmidrule(lr){5-6}
\textbf{Shape} & $\nu$ & RMSE &  & $\nu$ & RMSE  \\
\midrule
Logarithmic spiral, no outliers (Fig. \ref{fig:test_log_spiral}a) & - & 2.2$\cdot 10^{-3}$ &   & - & 2.2$\cdot 10^{-3}$ \\
 - with outliers, non-robust fitting (Fig. \ref{fig:test_log_spiral}b) & - & 9.5$\cdot 10^{-2}$ &   & - & 1.9$\cdot 10^{-1}$ \\
 - with outliers, robust fitting (Fig. \ref{fig:test_log_spiral}c) & 0.96 & 9.5$\cdot 10^{-2}$ &   & 0.99 & 9.5$\cdot 10^{-2}$\\
\midrule
Bent helix, no outliers (Fig. \ref{fig:test_bent_helix}a) & - &2.8$\cdot 10^{-1}$ &  & - & 3.9$\cdot 10^{-2}$ \\
 - with outliers, non-robust fitting (Fig. \ref{fig:test_bent_helix}b) & - & 3.2$\cdot 10^{-1}$  &   & - & 2.0$\cdot 10^{-1}$ \\
 - with outliers, robust fitting (Fig. \ref{fig:test_bent_helix}c) & 1.75 & 3.2$\cdot 10^{-1}$  &   & 1.30 & 1.6$\cdot 10^{-1}$\\
\bottomrule
\end{tabular}
\end{adjustbox}
\label{tab:validation}
\end{table*}


\subsection{Center Line and Core Line Extraction}
\begin{figure}[ht!]
\centering
\includegraphics[trim=60 20 80 0, clip,scale=.55]{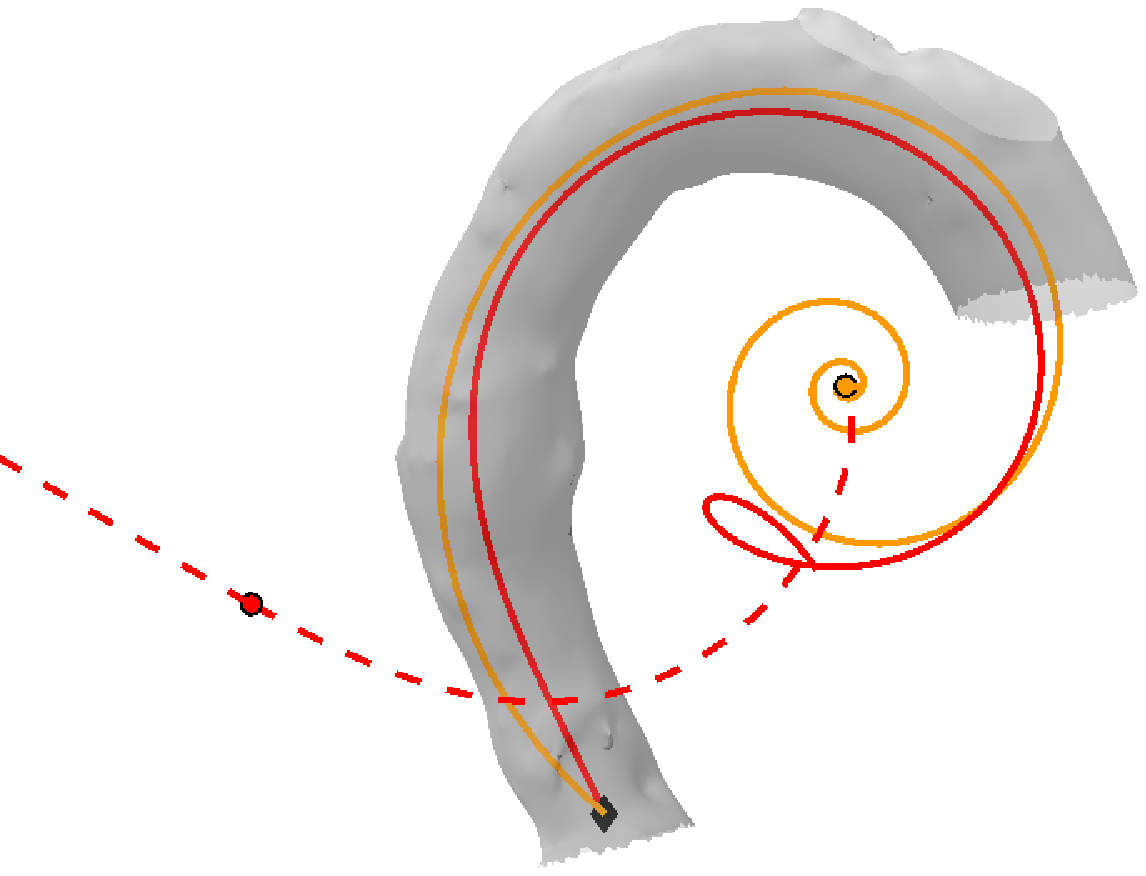}
\includegraphics[trim= 30 20 120 0, clip,scale=.55]{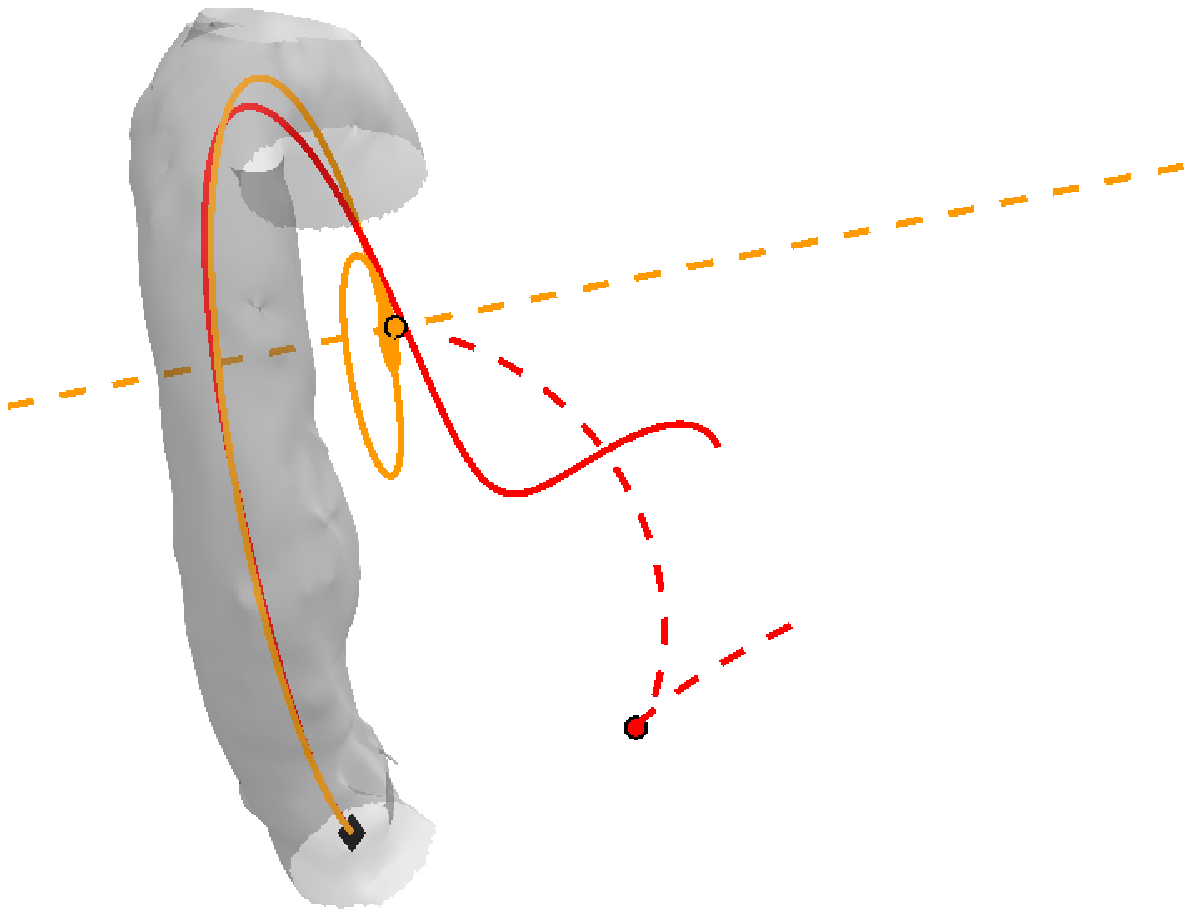}
\caption{Kinematic surface fitting of a human aortic arch \citep{onlineA}. Displayed are the results for the first order rotational (yellow curves) and second order rotational (red curves) velocity fields.
The dashed curves indicate the detected core lines, and convergence points $\pos_0$ are shown as circles.
Continuous curves represent example center lines $\curve$ traced from a central seed point $\pos_\mathrm{s}$ (black diamonds).}\label{fig:test_aorta}
\end{figure}

\begin{figure}[ht!]
\centering
\includegraphics[trim=100 15 100 00, clip,scale=.6,angle= 0]{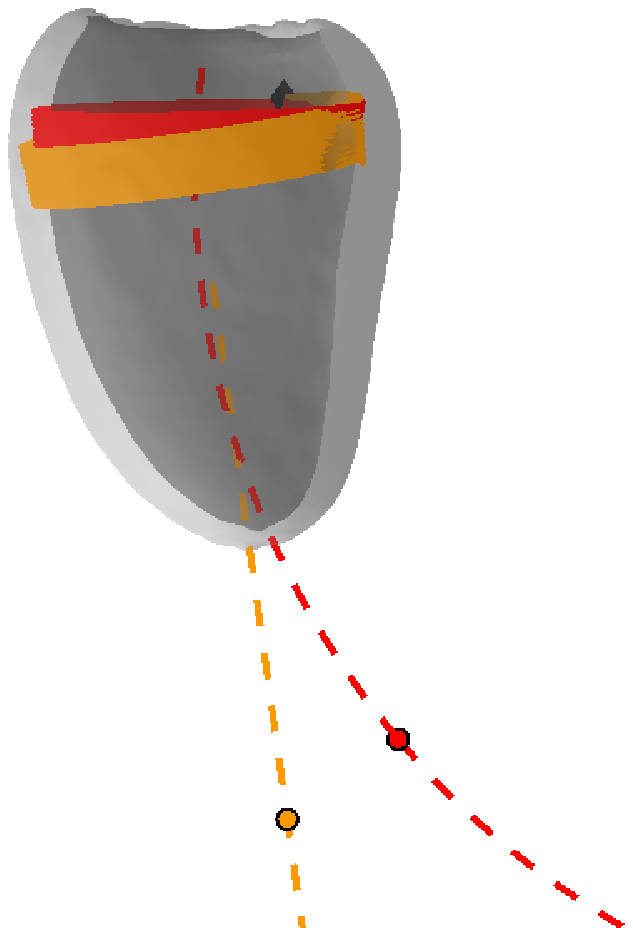}
\includegraphics[trim=100 0 100 50, clip,scale=.6,angle= 0]{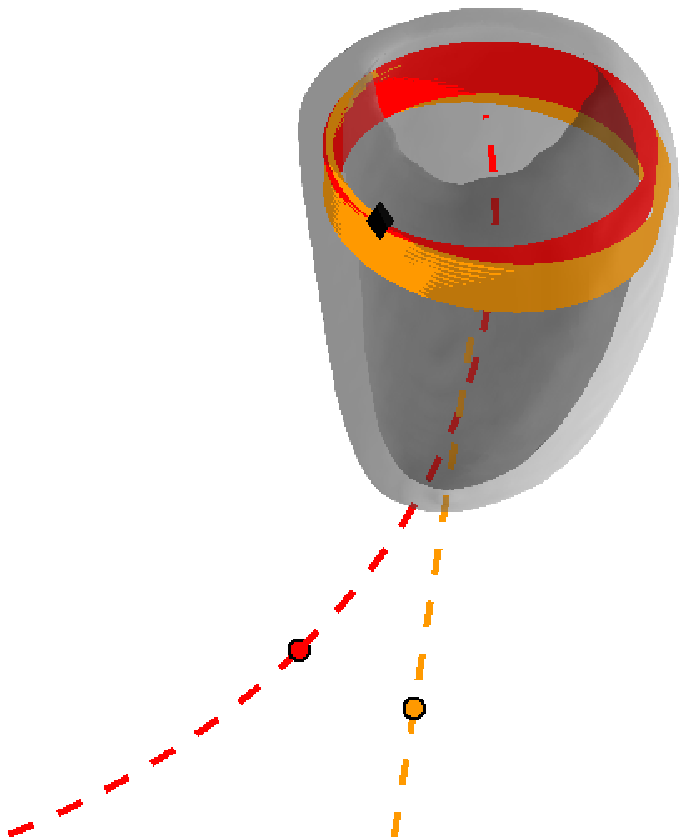}
\caption{Kinematic surface fitting of a human left ventricle \citep{bai2015bi}. Displayed are the results for the first order rotational (yellow curves) and second order rotational (red curves) velocity fields.
The dashed curves indicate the detected core lines, and convergence points $\pos_0$ are shown as circles.
Continuous curves represent example center lines $\curve$ traced from a central seed point $\pos_\mathrm{s}$ (black diamonds).} \label{fig:test_left_ventricle}
\end{figure}

\begin{figure}[ht!]
\centering
\includegraphics[trim=100 10 100 80, clip,scale=.6]{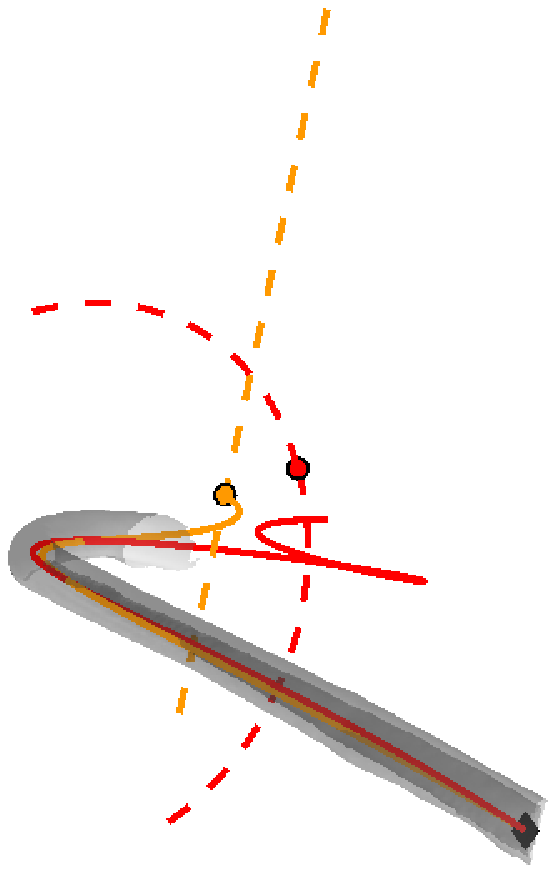}
\includegraphics[trim=100 30 80 80, clip,scale=.6]{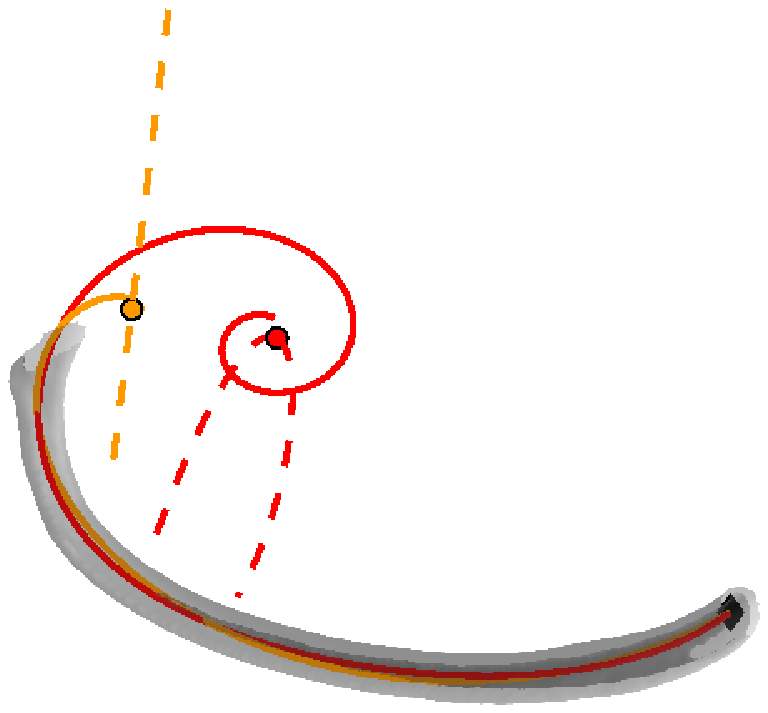}
\caption{Kinematic surface fitting of a human rib \citep{onlineR}. Displayed are the results for the first order rotational (yellow curves) and second order rotational (red curves) velocity fields.
The dashed curves indicate the detected core lines, and convergence points $\pos_0$ are shown as circles.
Continuous curves represent example center lines $\curve$ traced from a central seed point $\pos_\mathrm{s}$ (black diamonds)} \label{fig:test_rib} 
\end{figure}

\begin{figure}[!t]
\centering
\includegraphics[trim=80 40 40 40, clip,scale=.55]{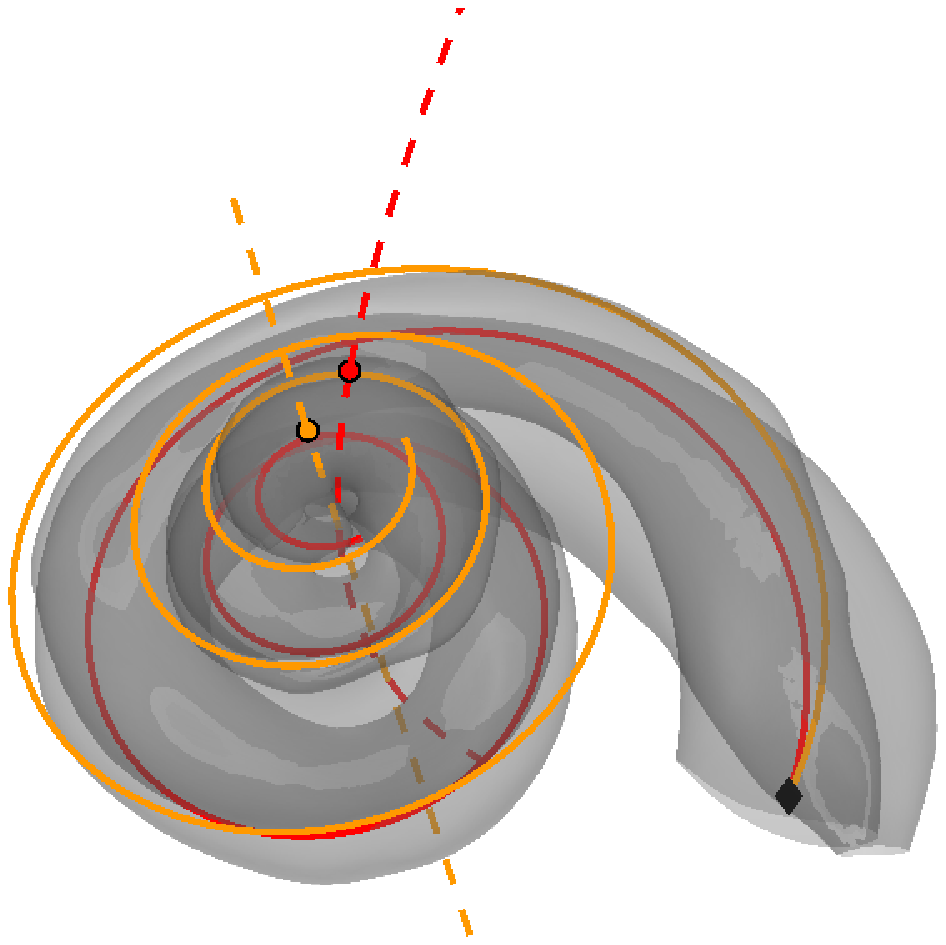}
\includegraphics[trim=80 40 40 60, clip,scale=.60]{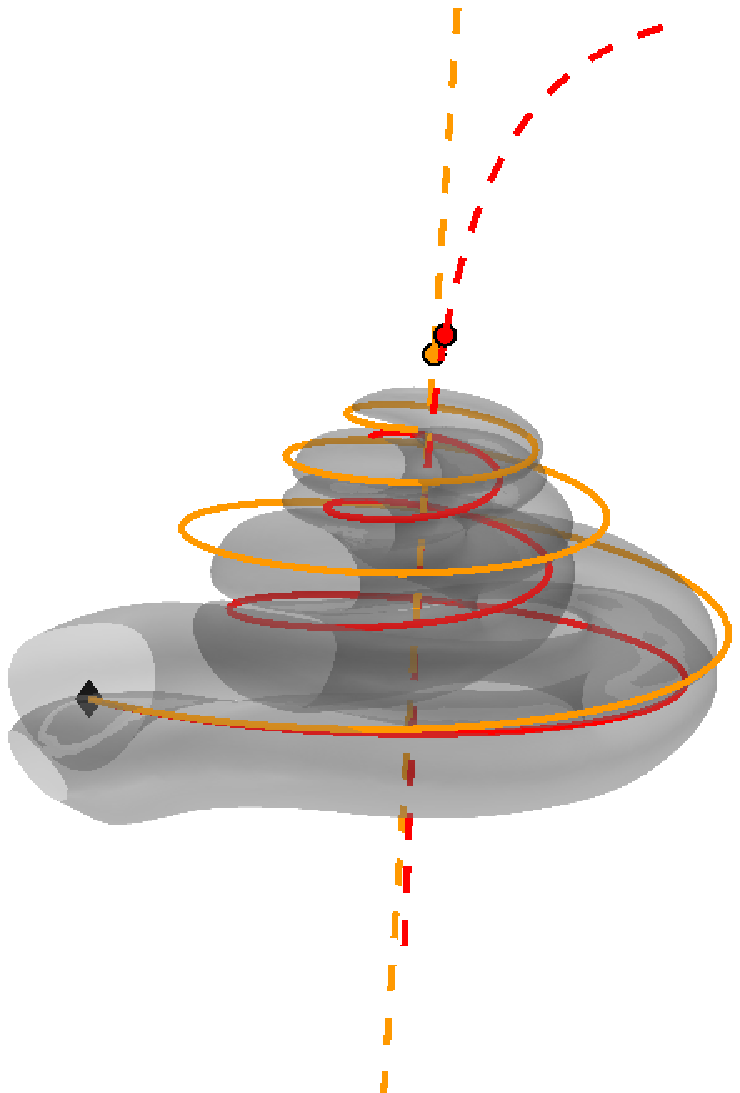}
\caption{Example of a right human cochlea \citep{sieber2019openear}. Displayed are the results for the first order  (yellow curves) and second order  (red curves) velocity fields.
The dashed curves indicate the detected core lines, and convergence points $\pos_0$ are shown as circles.
Continuous curves represent example center lines $\curve$ traced from a central seed point $\pos_\mathrm{s}$ (black diamonds). Note how the first order rotational scale velocity field (corresponding to the generation of a logarithmic spiral) fails to follow the twist between turns. The second order velocity field, however, can capture this degree of freedom.} \label{fig:test_cochlea}
\end{figure}

\begin{figure}[!t]
\centering
\includegraphics[trim=80  30 80  00, clip,scale=.55]{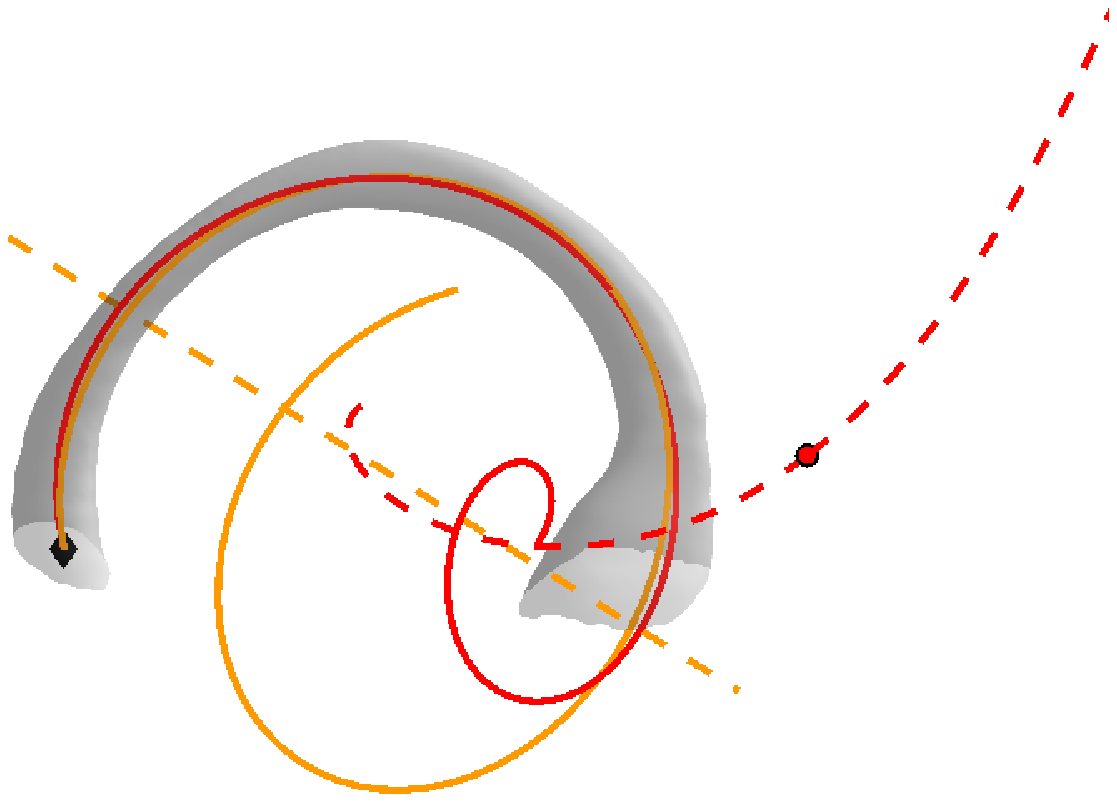}
\includegraphics[trim=80  50 80 20, clip,scale=.55]{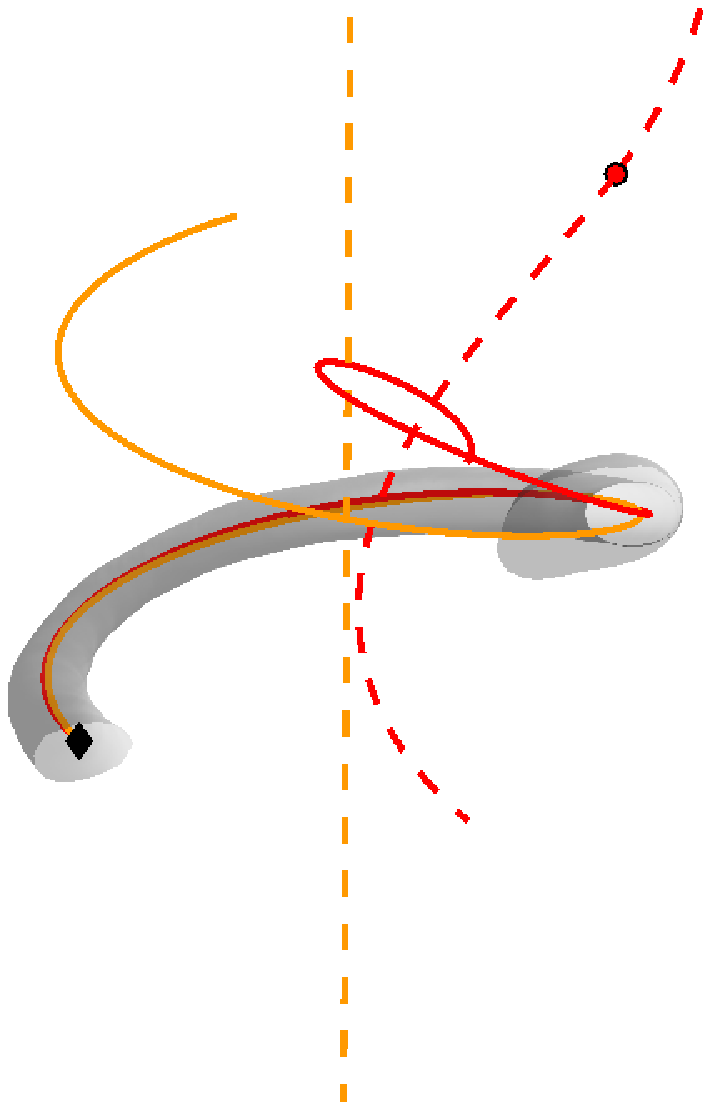}
\caption{Kinematic surface fitting of a human semicircular canal \citep{wimmer2019human}. Displayed are the results for the first order rotational (yellow curves) and second order rotational (red curves) velocity fields.
The dashed curves indicate the detected core lines, and convergence points $\pos_0$ are shown as circles.
Continuous curves represent example center lines $\curve$ traced from a central seed point $\pos_\mathrm{s}$ (black diamonds).} \label{fig:test_semi}
\end{figure}

Figures \ref{fig:test_aorta} to \ref{fig:test_semi} illustrate the extraction of center lines (streamlines extracted from a central seed point) and core lines in different human anatomical structures.
In all figures, we show the kinematic surface fitting results using the first order (yellow curves) and second order (red curves)   velocity fields.
The dashed curves indicate the detected  core lines for each velocity field. 
Continuous curves represent streamlines $\curve$ of the best fitting fields traced from selected central seed points $\pos_\mathrm{s}$ (black diamonds). 
The seed points were found by computing the centroid of vertices of a cross-section of the surface aligned with the rotational axis \citep{Hofer2005}.  
The parameters found for both velocity fields are summarized in Table \ref{tab:anatomical}.
 
\begin{table*}[htb!]
\centering
\caption{Kinematic surface fitting parameters for anatomical structures (15 iterations, $w_\pos = 0.001$).}
  \begin{adjustbox}{width=\textwidth}
\begin{tabular}{lcccccccccccccc}
\toprule
 & \multicolumn{6}{c}{\textbf{First order rotational scale velocity field}}  & & \multicolumn{4}{c}{\textbf{Second order rotational scale velocity field}}  \\
\cmidrule(lr){2-7}\cmidrule(lr){8-15}
\textbf{Anatomical structure} & $\rot$ & $\mathbf{c}$ & $\gamma$ & & $\nu$ & RMSE & & $\tors$ & $\rot$ & $\mathbf{c}$ & $\gamma$ & & $\nu$ & RMSE  \\
 \midrule
Aortic arch (Fig. \ref{fig:test_aorta}) & $[0.83,-0.43,-0.27]^T$& $[0.03,0.04,0.11]^T$& -0.20 & & 4.68 & 2.2$\cdot 10^{-1}$ & & $[0.05,0.76,0.46]^T$ & $[0.36,-0.17,-0.16]^T$& $[-0.04,0.07,0.06]^T$& -0.09 & & 5.52 & 1.9$\cdot 10^{-1}$\\
Left ventricle (Fig. \ref{fig:test_left_ventricle}) & $[0.02,-0.16,0.99]^T$& $[-0.02, 0.008, 0.003]^T$& -0.002 & & 6.89 & 1.2$\cdot 10^{-1}$ &  & $[0.24,-0.18,-0.03]^T$ & $[0.03,-0.14,0.98]^T$& $[0.001,-0.005,0.0004]^T$& -0.002 & & 8.77 & 1.1$\cdot 10^{-1}$\\
Rib (Fig. \ref{fig:test_rib}) & $[0.12,-0.11,-0.72]^T$& $[0.06,-0.33,0.24]^T$& -0.53 &  & 3.17 &  1.1$\cdot 10^{-1}$& & $[-0.37,-0.70,0.20]^T$ & $[0.06,-0.16,-0.43]^T$& $[0.02,-0.17,0.09]^T$& -0.27 & & 2.74 & 8.3$\cdot 10^{-2}$\\
Cochlea (Fig. \ref{fig:test_cochlea}) & $[-0.14,-0.09, 0.97]^T$& $[-0.09,0.13,0.03]^T$& -0.08 & & 2.51 & 1.8$\cdot 10^{-1}$ & & $[-0.46,0.23,0.17]^T$ & $[-0.14,-0.03,0.81]^T$& $[-0.10,0.12,0.02]^T$& -0.13 &  & 2.19 & 1.6$\cdot 10^{-1}$\\
Semicircular canal (Fig. \ref{fig:test_semi}) &   $[-0.53,0.76,-0.36]^T$ & $[-0.03,-0.13,0.01]^T$& -0.005 & & 1.84& 1.5$\cdot 10^{-1}$ & & $[-0.40,0.43,0.39]^T$& $[-0.47,0.42,-0.31]^T$& $[-0.08,-0.04,-0.04]^T$& -0.03 & & 1.80 & 1.7$\cdot 10^{-1}$\\
\bottomrule
\end{tabular} 
\end{adjustbox}
\label{tab:anatomical}
\end{table*}

\subsection{Morphological Classification}
\begin{figure}[ht!]
\centering
\includegraphics[trim=350 10 290 0, clip,scale=.25]{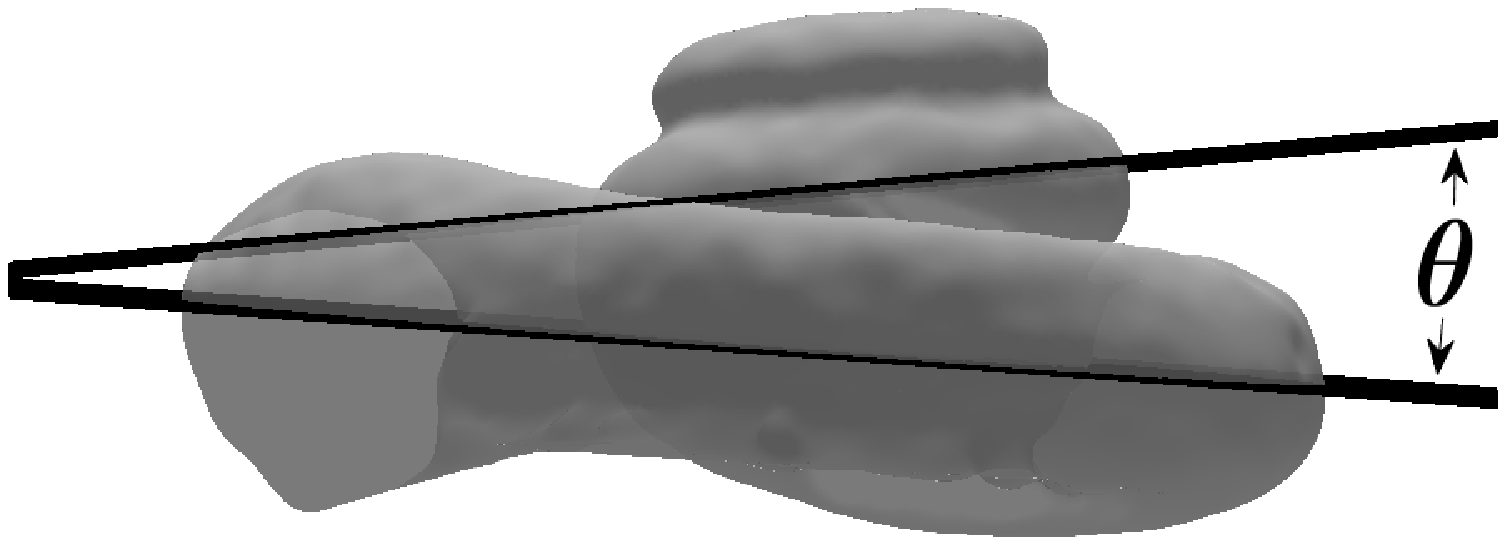}
\includegraphics[trim=50 5 50 10, clip,scale=.5]{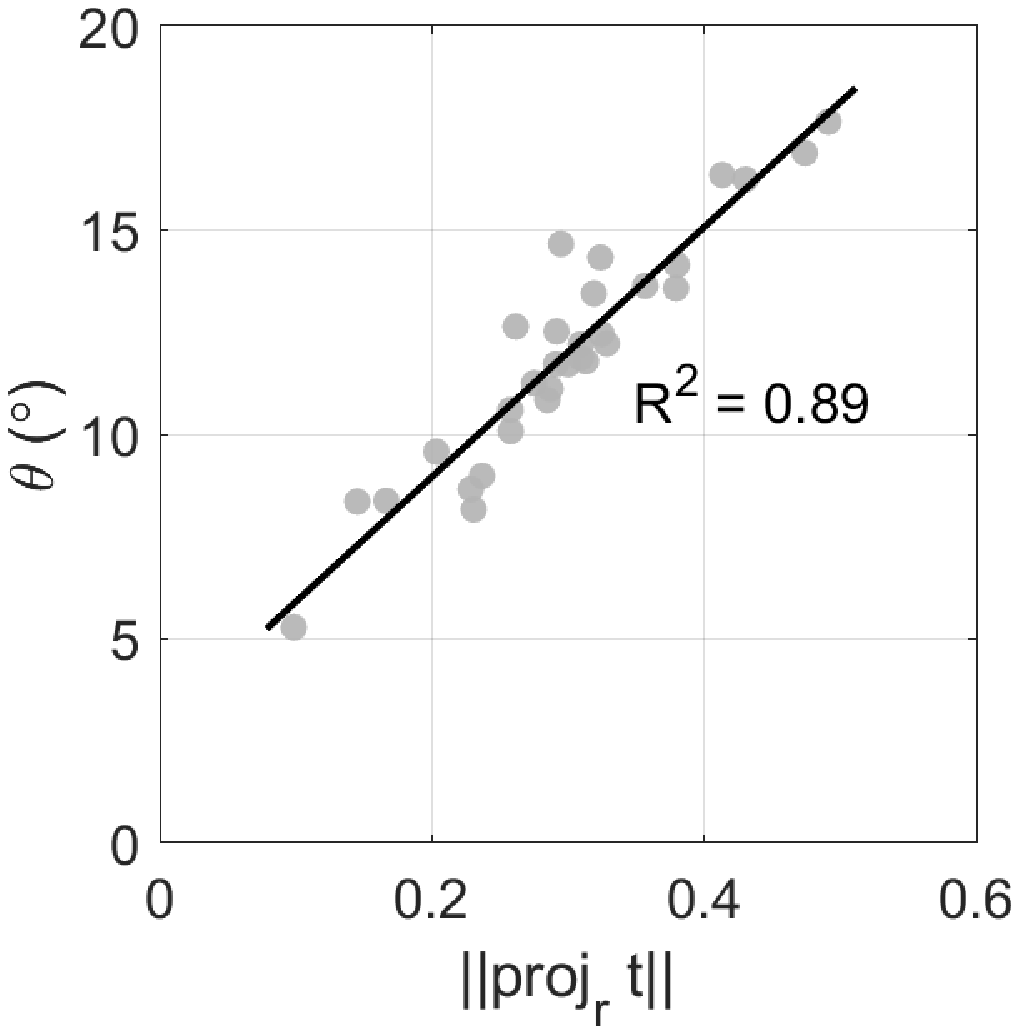}
\caption{Relationship between the intrinsic shape parameter $\| \mathrm{proj}_\rot \tors \| $ and the measured tilt angle $\theta$ 
\citep{shin2013quantitative} for 31 samples \citep{wimmer2019human,sieber2019openear}. The linear regression line and the coefficient of determination are shown.} \label{fig:test_class} 
\end{figure}

Kinematic surface fitting can be used to extract intrinsic global shape parameters for the morphological classification of anatomical structures. 
Here, we demonstrate an example of human cochlea classification.
The cochlea has a spiral shape with successive turns inclined towards each other, as depicted in Fig. \ref{fig:test_cochlea}. 
To quantify the tilt, we measure the angle $\theta$ between the basal turn and the middle turn of the cochlea, as defined by \citet{shin2013quantitative} (see Fig. \ref{fig:test_class}).
As an intrinsic shape parameter, we calculate the magnitude of the projection of $\tors$ on $\rot$, that is, $\| \mathrm{proj}_\rot \tors \|  = \frac{\tors \cdot \rot}{\|\rot\|}$.  
The correlation between the parameter $\|\mathrm{proj}_\rot \tors \|$ and the measured tilt angle is demonstrated in a set of 31 human cochleae \citep{wimmer2019human,sieber2019openear}.
This association could be used to classify a cochlea in terms of coiling and tilting characteristics, i.e., 'sloping' vs. 'rollercoaster' shapes \citep{avci2014variations}.

\section{Discussion}
We introduce a new approach for fitting kinematic surfaces to anatomical structures using a second order velocity field.
This advancement permits for the incorporation of more intricate shapes and enhances the precision of symmetry detection. 
Our method not only permits the recognition of curved rotational symmetries ({\em core lines}), but also facilitates morphological categorization by computing intrinsic shape parameters related to curvature and torsion.

The nature of the velocity field enables to identify rotational symmetries of first and second orders. 
For instance, in the human cochlea (Fig. \ref{fig:test_cochlea}), the highest order core line is the modiolar axis, while a generator curve traced out from the center of the duct represents the symmetry of the duct. 
Depending on the purpose, detection of a first order rotational symmetry may be sufficient, e.g., in the semicircular canal the detection of the first order rotational symmetry axis represents its orientation (Fig. \ref{fig:test_semi}). 
Similarly, a first order rotational field can be used as an approximation for symmetry detection, e.g., in a heart ventricle (Fig. \ref{fig:test_left_ventricle}) or for modiolar axis detection in cochleae as shown by \citet{wimmer2019}).

It is noteworthy that the recognition of revolutional or spiraling core lines is a frequent problem that arises when attempting to identify vortices (i.e., rotary motions) in the examination of fluid flows in various areas such as medicine, technology, meteorology, and more \citep{Gunther2018}. 
We could therefore envision applications of the technique in biomedical fluid mechanics to extract parametric core lines to define axial flow and link duct geometries to flow phenomena, which is relevant in the cardiovascular system and inner ear \citep{harte2023transverse,harte2023embc,stokes2023aneurysmal}.

Additional applications lie in segmentation and surface generation, where velocity fields can be used to trace out kinematic surfaces using generator curves \citep{Hofer2005}.
It is further possible to explore how useful our approach is for modeling the biological growth of tissue, e.g., bones \citep{xu2014construction}.

The global kinematic characterization of shapes may be desirable for the classification of suitable structures, e.g., the cochlea (Fig. \ref{fig:test_cochlea}). 
However, our method is limited in that it does not enable for a local or piecewise extraction of surface features, including multiscale representations. 
The algorithm should be adapted to be responsive to local variations in symmetry.
 
The introduced second order velocity field produces a class of kinematic surfaces that extends the possibilities of the first order field.
However, the number of described shapes is by no means exhaustive and limited to a certain set of solutions. 
For example, a toroidal helix can only be approximated by the velocity field. 
In a broader sense, we are not aware of a closed-form solution to detection core lines and convergence points.

\section{Conclusions}
This paper presents a novel approach for the fitting of kinematic surfaces to anatomical structures using a second order rotational scale velocity field. 
This advancement allows for the detection of more complex shapes, including those with curved rotational axes or core lines, enhancing the precision of symmetry detection and morphological classification. 
The method is capable of identifying rotational symmetries of both first and second orders, making it versatile for various applications in medical image analysis.

The results of our validation experiments demonstrate the effectiveness of the proposed approach, particularly in cases where first order velocity fields fail to capture the underlying shape characteristics. The introduction of a robust fitting scheme further enhances the method's resilience to outliers, making it suitable for real-world biomedical data.

By extending the capabilities of kinematic surface fitting, this approach opens up new possibilities for the characterization and classification of anatomical structures based on their intrinsic shape properties. 
It has the potential to be applied in various biomedical contexts, including the analysis of cardiovascular structures, the inner ear, and other complex biological systems.
Our approach could be of use in radiology, orthopedics, and neurology, where anatomical shape examination and symmetry core line detection is necessary for diagnosis, treatment preparation, and surgical procedure planning.
There are potential uses of biomedical fluid mechanics that are worth exploring, as well as the application to model tissue growth, e.g., bone formation.

\section*{CRediT Authorship Contribution Statement}
\textbf{Wilhelm Wimmer:} Conceptualization, Methodology, Software, Validation, Formal analysis, Investigation, Writing – original draft, Writing – review \& editing, Visualization, Funding acquisition. \textbf{Herv\'{e} Delingette:} Conceptualization, Methodology, Formal analysis, Writing – original draft, Writing – review \& editing.

\section*{Acknowledgments}
This work was supported by the Swiss National Science Foundation under project grant number 205321\_200850 and by the French government, through the 3IA Côte d’Azur Investments in the Future project managed by the National Research Agency (ANR)  with the reference number ANR-19-P3IA-0002.

 
\end{document}